\documentclass[11pt]{article}
\usepackage{geometry}
\usepackage{coling2020}
\usepackage{times}
\usepackage{url}
\usepackage{latexsym}
\usepackage{microtype}
\usepackage{amsmath}
\usepackage{amssymb}
\usepackage{csquotes}
\usepackage{mathtools}
\usepackage{graphicx}
\usepackage{subcaption}
\usepackage{rotating}
\usepackage{MnSymbol,wasysym}

\colingfinalcopy 

\title{ULD@NUIG at SemEval-2020 Task 9: Generative Morphemes with an Attention Model for Sentiment Analysis in Code-Mixed Text}

\author{\begin{tabular}{c} Koustava Goswami, Priya Rani, Bharathi Raja Chakravarthi, \\ Theodorus Fransen, and John P. McCrae \end{tabular}\\
Insight SFI Research Centre for Data Analytics,\\
Data Science Institute, National University of Ireland Galway\\ \tt \{koustava.goswami, priya.rani,  bharathi.raja, \\ \tt theodorus.fransen, john.mccrae\}@insight-centre.org\\}

\date{}

\begin{document}

\maketitle

\begin{abstract}

Code mixing is a common phenomena in multilingual societies where people switch from one language to another for various reasons. 
 Recent advances in public communication over different social media sites have led to an increase in the frequency of code-mixed usage in written language. In this paper, we present the Generative Morphemes with Attention (GenMA) Model sentiment analysis system contributed to SemEval 2020 Task 9 SentiMix. The system aims to predict the sentiments of the given English-Hindi code-mixed tweets without using word-level language tags instead inferring this automatically using a morphological model. The system is based on a novel deep neural network (DNN) architecture, which has outperformed the baseline F1-score on the test data-set as well as the validation data-set. Our results can be found under the user name \enquote{koustava} on the \enquote{Sentimix Hindi English}\footnote{\url{https://competitions.codalab.org/competitions/20654#learn_the_details-results}} page.
\end{abstract}

\section{Introduction}

%
%
\blfootnote{
    %
    %
    %
    %
    \hspace{-0.65cm}  
    This work is licensed under a Creative Commons 
    Attribution 4.0 International Licence.
    Licence details:
    \url{http://creativecommons.org/licenses/by/4.0/}.
    %
    %
}

Sentiment analysis refers to a process of predicting the emotion content from a given text. Sentiment analysis is usually seen as a categorization problem over a variable with three values: positive, negative, neutral \cite{phani-etal-2016-sentiment}. With the increase in the popularity of social media such as Twitter, a new area of study to the field of natural language processing and thus, sentiment analysis has been explored. Most of the data extracted from social media are code-mixed \cite{7918035,9074379}, which have become a common approach in most cases but also pose unique challenges.  

Analysis of short texts from micro-blogging platforms such as Twitter is in high demand as the analysis of these text distill and evaluate the moods and the sentiment of the users and are very useful for different organisations, be it government or business or NGO. Sentiment analysis for Indian code-mixed languages is relatively new \cite{9074205,chakravarthi-etal-2020-sentiment,chakravarthi-etal-2020-corpus,rani-etal-2020-comparative}. The significant difference in style of language, orthography \cite{chakravarthi-et-alortho} and grammar used in tweets presents specific challenges for English-Hindi code-mixed data. In this paper we aim to introduce a novel deep neural network system which was submitted for SemEval 2020 Task 9, Sub Task A for English-Hindi data \cite{patwa2020sentimix}. We will also compare the system with other state-of-the-art systems and describe how the system has outperformed others. The systems were trained using only the Twitter data provided by the organisers excluding the word-level language tags provided in the data. 

\section{Related Work} \label{rel}

Although the field of sentiment analysis is growing and several systems have advanced the state-of-the-art, the overall performance of systems to predict sentiment in code-mixed data is low. \newcite{7275819} predicted overall sentiment score for Hindi-English code-mixed data using a lexicon based approach. \newcite{go2009twitter} were the first to look at the task as a query-driven classification problem. A Hindi-English data-set was introduce by \newcite{joshi-etal-2016-towards} for sentiment analysis and they performed empirical analysis comparing the performance of various state of the art models in sentiment analysis. They also introduced a sub-word level representation in an LSTM model instead of character or word level representation. \newcite{Dos-santos-gatti-2014-deep} proposed a deep convolutional neural network that exploits character level and sentence level information to predict the sentiments in short texts. All these previous experiments were dependent on the word-level language tags, and this is a disadvantage as it is time-consuming to annotate at the word level. In our approach, we create a model without the need for word-level annotation.

\section{Dataset}\label{da}
The dataset used for the current task is provided by SentiMix English-Hindi Task 9 in SemEval-2020 \cite{patwa2020sentimix}. It consists of English-Hindi code-mixed tweets annotated with sentiment labels: positive, negative, or neutral. Besides the sentiment labels the data-set also includes word-level language tags, which are \textit{en} (English), \textit{hi} (Hindi), \textit{mixed}, and \textit{univ} (symbols, @ mentions, hashtags). As it is very common for Twitter data to have other forms of text such as URLs and emoticons, this data-set too contains emojis such as \textbf{\smiley{}} \textbf{\frownie{}} and URLs.


The pre-processing removes the word-level language tags. 
We normalize the data for training the Support Vector Machine (SVM) and deep neural network (DNN), by lower-casing all the tweets and removing punctuation, emojis and URLs. After converting all the tweets into lower case, extra spaces were removed from the tweets. The tweets are tokenized into characters, where each character has been mapped to an index number. The character-index mapping is created with the help of the Keras tokenizer package\footnote{\url{https://www.tensorflow.org/api_docs/python/tf/keras/preprocessing/text/Tokenizer}}.

\section{System Description}\label{des}
\subsection{Support Vector Machine}
The Support Vector Machine (SVM) is an algorithm which maximizes a particular mathematical function with respect to a given collection of data \cite{noble2006support}. In our experiment, we have focused on the linear SVM methodology. The objective of linear SVM optimization problem is to maximize the given equation:
            \begin{equation}
                max_{\alpha}\sum_{i=1}^{l}\alpha_{i}-\frac{1}{2}\sum_{i=1}^{l}\sum_{j=1}^{l}y_{i}y_{j}\alpha_{i}\alpha_{j}(x_{i}x_{j})
            \end{equation}
where $\alpha_{i}$ is the weight of the examples, \textit{x} is the input and \textit{y} is the label. After pre-processing the data, we experimented with the most basic input feature TF-IDF and was created with the help of TfidfVectorizer\footnote{\url{https://scikit-learn.org/stable/modules/generated/sklearn.feature_extraction.text.TfidfVectorizer.html}} of the Scikit Learn package. 

\subsection{Convolution Neural Network (CNN)}
In this experiment, we have followed 
the CNN model described by \newcite{zhang2015character} 
which has a one-character embedding layer and four convolution (CONV1D) layers. For the first three convolution layers, after each layer, one max-pooling layer has been added. 
In the end, one hidden layer is followed by one softmax layer. The model accepts sentences as sequence and characters as input. The character embedding is a one-hot embedding (1-to-n embedding) where the number of unique characters is \textit{n}. The shape of the filter is 1-dimensional of size \textit{k}. The filter slides over the input sequence matrix to create the feature map of dimension $b \times f \times s$ where \textit{b} is the batch size, \textit{f} is the number of filters used, and \textit{s} is determined by the formula $m -k +1$  where \textit{m} is the input size. Stride 1 is used to calculate features based on each character including spaces and special characters.

\subsection{Generative Morphemes with Attention (GenMA) Model}
\begin{figure}[!htbp]
    \centering
    \includegraphics[scale=0.35]{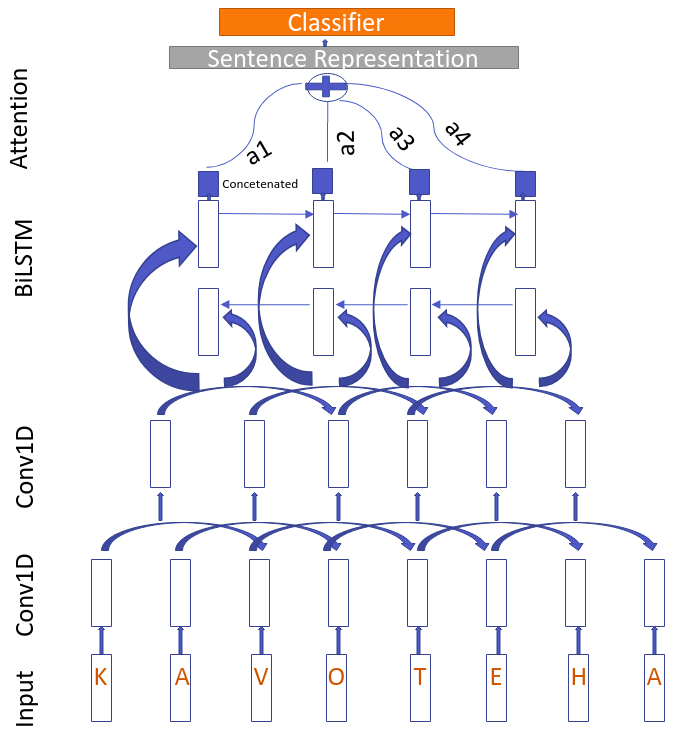}
    \caption{Relative Character Attention Model; the input is the character embeddings which is mentioned with physical characters; boxes filled with blue colour are the concatenated outputs of the BiLSTM. }
    \label{fig:char_att_model}
\end{figure}

We propose an Artificial Morphemes Generative system with Self Attention (SA) layer. The model takes the input sequence as a character sequence. The model has one character embedding layer and two convolution (CONV1D) layers. Each convolution layer has one max-pooling layer each.  After the convolution layers, there is one bidirectional LSTM layer followed by one self-attention layer. The model has one hidden layer and one softmax layer. 
\newcite{8411809} designed a network which is based on the Chinese character attention model for sentence classification. We have taken inspiration from that model. \newcite{8411809} rely on capturing local context of sentences based on a single convolution layer whereas  GenMA is capable of generating new artificial morphemes and framing a sentence as a group of new morphemes irrespective of language identification of source words. The combination of two CNN layers helps to generate new morphemes based on deep relative co-occurring characters (3 characters frame), and the LSTM layer helps to capture global information of sentences based on newly generated features. The SA layer helps to construct sentence-level information. It also captures relativity strength among different co-occurring character features. The model architecture can be found in Figure \ref{fig:char_att_model}. 

\textbf{A Convolution Neural Network } layer is used as a feature extractor of the sentences. The one-dimensional convolution implements 1-dimensional filters which slides over the sentences as a feature extractor. Let the filters have a shape of $1 \times k$ where \textit{k} is the filter size. Let $x_i\in\left\{\mbox{0,1}\right\}^n$ denote the one-hot representation of the \textit{i}-th character considering character vocabulary size is \textit{n}. For each position \textit{j} in the sentence, we have a window vector $w_j$ with \textit{k} consecutive character vectors \cite{zhou2015c} denoted as
\begin{equation}\label{equation-2}
    w_j=[x_j,x_{j+1},.....,x_{j+k-1}]
\end{equation}
The 1-dimensional \textit{k}-sized filters slide over the window vector $w_j$ to create the feature map \textit{s} where $ s\in\mathbb{R}^{m-k+1} $ and where \textit{m} is the input size. Multiple filters are used to generate different feature maps for each window $w_j$. The new feature representation, $W_j$, will represent a new feature map vector for the \textit{j}-th position of the sentence. The second convolution layer will take feature representations as input and generate a high-order feature representation of the characters. The max-pooling network after each convolution network helps to capture the most important features of size \textit{d}. The new high-order representations are then feed to the LSTM (Long Short Term Memory Network) as input.  

\textbf{Long Short Term Memory (LSTM) Network} layer takes the output of the previous CNN layer as input. It produces a new representation sequences in the form of $h_1,h_2,....h_n$ where $h_t$ is the hidden state of the LSTM of time step \textit{t}, summarising all the information of the input features (morphemes) of the sentences.
An LSTM unit is composed of one memory cell and three gates (input gate, forget gate and output gate) \cite{hochreiter1997long}. At each time step \textit{t}, the hidden state takes the previous time step hidden state $h_{t-1}$ and characters ($x_t$) as input. Let us denote memory cell, input gate, forget gate and output gate as $c_t$,$i_t$,$f_t$,$o_t$. The output hidden state $h_t$ and the memory cell $c_t$ of timestep \textit{t} is defined by Equation \ref{lstm}

    %
\begin{equation}\label{lstm}
    \begin{split}
    i_t = \sigma(W_i \cdot [h_{t-1},x_t] + b_i) \quad\text{,}\quad f_t = \sigma(W_f \cdot [h_{t-1},x_t] + b_f)\\
    o_t = \sigma(W_o \cdot [h_{t-1},x_t] + b_o) \quad\text{,}\quad f_t = \sigma(W_f \cdot [h_{t-1},x_t] + b_f)\\
    c_t = f_t \odot c_{t-1} + i_t \odot q_t \quad\text{,}\quad
    h_t = o_t \odot \tanh(c_t)
    \end{split}
\end{equation}

Here $\odot$ is the element wise operation, $W_i$,$W_f$,$W_o$,$W_q$ are the weights of the matrices, $b_i$,$b_f$,$b_o$,$b_q$ are the biases and $\sigma$ denotes the logistic sigmoid function. A Bidirectional LSTM (BiLSTM) network has been used which has helped us to summarise the information of the features from both directions. The Bidirectional LSTM consists of a forward and backward pass which gives us two annotations of the hidden state $h_{for}$ and $h_{back}$. We obtained the final hidden state representation by concatenating both the hidden states $h_i = h_{i-for} \oplus h_{i-back}$, 
where $h_i$ is the hidden state of the \textit{i}-th timestep and $\oplus$ is the element-wise sum between the matrices.

\textbf{The Attention} layer helps us to determine the importance of one morpheme over others while building sentence embeddings for classification. 
A self-attention mechanism has been adopted from \newcite{baziotis-etal-2018-ntua-slp-semeval} which will help to identify the morphemes that are important for capturing the sentiment of the sentence. The self-attention mechanism is built upon attention mechanism by \newcite{Bahdanau2014NeuralMT}. The attention mechanism assigns weight $a_i$ to each feature's annotation based on output $h_i$ of the LSTM's hidden states, with the help of the softmax function  as illustrated in Equation \ref{attention}
\begin{equation}\label{attention}
    \begin{split}
    a_i = \tanh(W_h\cdot h_i + b_h)
    \quad\text{,}\quad
    a_i = \frac{exp(a_i)}{\sum_{t=1}^{T}exp(a_t))}
    \end{split}
\end{equation}
The new representation will give a fixed representation of the sentence by taking the weighted sum of all feature label annotations as shown in Equation \ref{repre}
\begin{equation}\label{repre}
    r = \sum_{i=1}^{T}a_i \cdot h_i
\end{equation}
where $W_h$ and $b_h$ are the attention weights and bias respectively. 

\textbf{The Output} layer consists of one fully-connected layer with one softmax layer. The sentence representation after the attention layer is the input for the dense layer. The output of the dense layer is the input of the softmax which gives the probability distribution of all the classes with the help of the softmax function as shown in Equation \ref{soft}
\begin{equation}\label{soft}
    p_i = \frac{exp(a_i)}{\sum_{t=1}^{T}exp(a_t)}
\end{equation}
where $a_i$ is the output of the dense layer.

\subsection{Parameters}
 A Linear SVM kernel is used for the first SVM model. Other parameters are kept as mentioned in the implementation \cite{noble2006support}. 
 For the next two models, the convolution network setup is the same. We have used 32 filters and the kernel size is 3. The maxpooling size is 3. The hidden size $h_i$ of LSTM units is kept to 100. The dense layer has 32 neurons and it has 50 percent dropout. The Adam optimizer \cite{Kingma2015AdamAM} is used to train our model with the default learning set to 0.0001. The batch size is set to 10. For the convolution layer in both the experiments we have used the relu activation function \cite{nair2010rectified} and for the dense layer we have used tanh activation function \cite{kalman1992tanh}. Categorical cross entropy loss is used for the multi-class classification. We have used Keras\footnote{\url{https://keras.io}} to train and test our model.

\section{Results} \label{res}
Overall we see varying performance across the classifier, with some performing much better out-of-sample than others. Table \ref{table:2} shows the class-wise macro F1-score of the models on the test set for different models.
\begin{table}[!htbp]
\centering
\begin{tabular}{|l|l|l|l|l|}
\hline
\textbf{Model}    & \textbf{Pos Class} & \textbf{Neg Class} & \textbf{Neut Class} & \textbf{Score} \\ \hline
\textit{SVM}      & 0.64               & 0.62               & 0.57                & 0.61           \\ \hline
\textit{Char-CNN} & 0.68               & 0.65               & 0.56                & 0.63           \\ \hline
\textit{GenMA}    & 0.73               & 0.67               & 0.63                & \textbf{0.68}  \\ \hline
\end{tabular}
\caption{F1-scores of three algorithms on dataset}
\label{table:2}
\end{table}

The state-of-the-art character CNN model has performed better than the SVM model. One of the main reasons for a CNN to perform better than SVM is that a CNN is capable of identifying the features of the sentence with the help of neural model weight distribution. It also takes special characters into account which make the sentence embedding more robust to work on. On the other hand, the hyper-tuning settings of the tf-idf vectors could be the cause of lower performance of the SVM.

Our GenMA model has outperformed all classical models as well as the state-of-the-art character CNN model as it considers a sentence composed of a different set of morphemes. The individual results on three different sentiment classes show that the model outperforms the other two models while recognizing individual classes whereas the SVM model recognizes neutral classes better than the CNN model. Our model has achieved 0.68 F1-score in the test set which is 7 percent better than the SVM and 5 percent better than the character CNN model. 

\section{Discussion} \label{desc}
Our proposed GenMA model outperforms other models as it is capable of generating new morphemes out of neighbor characters and it identifies the essential morphemes to classify a sentence. 
The two main advantages of our model are: 
\begin{itemize}
    \item The model can construct sentence embeddings based on the new generative morphemes which are created artificially in combination of both the languages Hindi and English. These morphemes carry the features of both the languages.  As illustrated in figure \ref{fig:3}, the new morpheme \textbf{avo} generated by the model, where the character \enquote{a} is taken from the Hindi word \enquote{ka} and the character \enquote{vo} belongs to the English word \enquote{vote}, shows that these new artificial generative morphemes have features of both Hindi and English. Thus, the multilingual word-level language identification annotations are not required.
    \item  The model is able to correctly identify the co-occurring character sets with highest importance in sentiment analysis. The attention mechanism is visualized in Figure \ref{fig:3}. 
    \begin{figure}[h]
    \centering
    \includegraphics[scale=0.30]{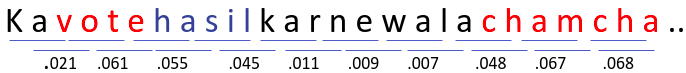}
    \caption{Character of tweets (English-Hindi) with attention}
    \label{fig:3}
\end{figure}
The red characters are the important characters followed by blue characters. The black characters are contributing least significantly to the sentence classification. In the generated artificial morpheme, some morphemes put more emphasis on sentence polarity (An example is morpheme \enquote{ote}, which weights 5 times (0.061) than the normal morpheme \enquote{arn} (0.011)). The softmax attention weights are able to rank character importance from high to low. 
\end{itemize}

\section{Conclusion} \label{con}
In this paper we have proposed a novel deep neural model which has outperformed the baseline scores on code-mixed data proposed in \newcite{patwa2020sentimix} and state-of-the-art models discussed in Section \ref{res}. Our model is capable of classifying the sentiment of the sentences without considering language difference between words in the sentences with an F1-score of 0.68 on the test data. 

Future work may reveal how to capture sentiment based on emojis that are widely used in tweets. One of our settings is artificial morpheme generation for the Hindi and English dataset. But we have not explored this method in the context of morphologically complex code-mixed datasets. We will aim to implement the model in the complex code-mixed dataset in the future. We will also try to capture word level information of code-mixed sentences without language identity to understand what the important key words are to classify sentences. 

\section*{Acknowledgments}
This publication has emanated from research in part supported by the Irish Research Council under grant number IRCLA/2017/129 (CARDAMOM-Comparative Deep Models of Language for Minority and Historical Languages) and co-funded by Science Foundation Ireland (SFI) under Grant Number SFI/12/RC/2289 (Insight),  SFI/12/RC/2289$\_$P2 (Insight$\_$2).
\bibliographystyle{coling}
\bibliography{semeval2020}

\end{document}